\documentclass[letterpaper, 10 pt, conference]{ieeeconf}
\IEEEoverridecommandlockouts
\usepackage[utf8]{inputenc}
\usepackage[noend]{algpseudocode}
\usepackage{amsmath}
\usepackage{amssymb}  
\usepackage{xcolor}
\usepackage{graphicx}
\usepackage{hyperref}
\usepackage{booktabs}
\usepackage{color}
\usepackage{array}
\usepackage{multirow}
\usepackage{balance}
\usepackage{siunitx}
\usepackage{tabularx}
\usepackage{makecell}
\usepackage{cite}
\usepackage[linesnumbered,boxed,ruled,commentsnumbered]{algorithm2e}
\graphicspath{{figures/}}
\DeclareGraphicsExtensions{.pdf,.png,.jpg,.eps}

\begin{document}

\title{\LARGE \bf FAST-LIVGO: A Degeneracy-Robust LiDAR-Inertial-Visual-GNSS Fusion Odometry}

\author{Zhiyu Chen$^{1*}$, Chunran Zheng$^{2*}$, Jiayu Wen$^{1}$, Xiaolei Zhang$^{1}$, Jiaming Xu$^{1}$, Feng Pan$^{3}$, and Yukang Cui$^{1\dagger}$
\thanks{*These authors contributed equally to this work.}%
\thanks{$^{\dagger}$Corresponding author.}%
\thanks{This work was partially supported by National Natural Science Foundation of China (62573303, 61903258), Guangdong Basic and Applied Basic Research Foundation (2024A1515030153), and the Project of Department of Education of Guangdong Province (2023ZDZX4046).}%
\thanks{$^{1}$Zhiyu Chen, Jiayu Wen, Xiaolei Zhang, Jiaming Xu, and Yukang Cui are with the College of Mechatronics and Control Engineering, Shenzhen University, Shenzhen 518060, China. (e-mail: cuiyukang@gmail.com)}%
\thanks{$^{2}$Chunran Zheng is with the Department of Mechanical Engineering, The University of Hong Kong, Hong Kong SAR, China. (e-mail: zhengcr@connect.hku.hk)}%
\thanks{$^{3}$Feng Pan is with the College of Automation, Harbin Engineering University, Harbin 150001, China. (e-mail: fengpan97618@163.com)}%
}
\maketitle
\thispagestyle{empty}

\begin{abstract}
Robust state estimation and mapping in long-term, large-scale, and highly dynamic environments remains a key challenge in robotics. Existing LiDAR-Inertial-Visual Odometry (LIVO) systems achieve strong local accuracy but suffer from accumulated drift over long distances and may fail in geometrically degraded or textureless scenes. Meanwhile, GNSS-aided fusion frameworks often rely on LiDAR or visual odometry for state prediction and outlier rejection, making them vulnerable when odometry degenerates. To address these limitations, we propose a tightly coupled LiDAR-Inertial-Visual-GNSS fusion framework based on an Error-State Iterated Kalman Filter. An online spatiotemporal alignment module using Dynamic Time Warping is introduced for highly dynamic conditions. To better exploit GNSS precision, we develop observation models based on Doppler shifts and fixed-anchor Time-Differenced Carrier Phase, providing millimeter-level relative constraints without augmenting historical anchor states. We further design a degeneracy-aware dual-mode outlier rejection strategy that switches between LIVO-prior-guided rejection and GNSS-aided recovery according to the LIVO degeneracy level. Experiments on the public M3DGR dataset and a custom 20~m/s fixed-wing UAV dataset demonstrate that our system reduces accumulated drift and map ghosting, outperforming state-of-the-art methods in accuracy and robustness.
\end{abstract}
	
	\IEEEpeerreviewmaketitle
	
    \section{INTRODUCTION}
    \label{sec:intro}
    
Achieving robust state estimation and environmental mapping in complex, large-scale, and highly dynamic environments remains a core challenge for autonomous systems \cite{cui2026}. To overcome single-sensor limitations, multi-sensor fusion, especially LiDAR-Inertial-Visual Odometry (LIVO), has been widely adopted by combining LiDAR's geometric accuracy with the camera's robustness in texture-rich areas \cite{camvox, fastlivo2, r3live, lvisam, vinsmono}, while recent RGB--thermal fusion further improves mapping robustness in low-light environments \cite{fast_rtfm}. However, in large-scale field missions or high-speed Unmanned Aerial Vehicle (UAV) flights, pure LIVO systems still face critical limitations. The absence of global constraints leads to cumulative drift that grows over time and distance, often diverging along the Z-axis. Moreover, LIVO may fail in perceptual degradation scenarios, such as textureless open squares, featureless walls, or high-speed motion blur.

Although Global Navigation Satellite Systems (GNSS) can provide absolute positioning to suppress LIVO drift, tightly integrating GNSS remains challenging. In complex environments such as urban canyons or tree-lined avenues, Non-Line-of-Sight (NLOS) signals and multipath effects introduce severe anomalous noise, and directly fusing such contaminated measurements can degrade trajectory smoothness \cite{liosam, vinsfusion}. In addition, time synchronization errors of low-cost sensors are significantly amplified under high-speed motion, causing estimation lags. Existing loosely-coupled schemes underutilize the high precision of GNSS carrier phases, while traditional tightly-coupled frameworks lack adaptive integrity monitoring under alternating LIVO and GNSS degradation.

To address these issues, we propose a tightly-coupled multi-sensor fusion system formulated within an Error-State Iterated Kalman Filter (ESIKF), aiming at globally consistent mapping in highly dynamic scenarios and robust operation in severely degraded environments. The system exploits sensor complementarities by using high-frequency FAST-LIVO2 odometry as geometric priors for GNSS signal evaluation and online spatiotemporal alignment, while leveraging GNSS Doppler shifts and Time-Differenced Carrier Phase (TDCP) observations to provide absolute velocity and relative displacement constraints. Furthermore, a hierarchical integrity monitoring and adaptive fusion strategy is designed to reject GNSS multipath anomalies and support seamless indoor-outdoor transitions under LIVO geometric or textural degradation.

The main contributions are summarized as follows:
\begin{itemize}
    \item A tightly-coupled LiDAR-Inertial-Visual-GNSS fusion framework based on ESIKF. It handles spatiotemporal synchronization of low-cost sensors under highly dynamic conditions using Dynamic Time Warping (DTW) and a two-step extrinsic optimization. Precise Doppler and fixed-anchor TDCP observation models are further constructed to introduce sub-millimeter relative constraints without state augmentation, improving global consistency.
    
    \item A degeneracy-dependent dual-mode outlier rejection strategy. By analyzing the eigenvalues of the LIVO Hessian matrix for real-time environmental degradation detection, the system adaptively switches between LIVO-prior-guided rejection and GNSS-aided recovery, ensuring robust operation in both feature-sparse and satellite-occluded environments.
    
    \item Extensive validations on the public M3DGR dataset \cite{m3dgr} and our custom 20~m/s high-speed UAV and handheld datasets. Results demonstrate that the proposed system outperforms SOTA methods in localization accuracy and mapping consistency. The highly dynamic datasets will be open-sourced to support future research.
\end{itemize}
	
    \section{RELATED WORKS}
    \label{sec:related_works}
    
    \subsection{LiDAR-Inertial-Visual Odometry (LIVO)}
    \label{subsec:livo_related}
    Multi-sensor fusion is fundamental for robust state estimation. Early LIVO systems mainly adopted loose fusion, while R2LIVE \cite{r2live} advanced tight coupling by fusing LIO states within an ESIKF and processing visual reprojection errors in a parallel optimization thread. R3LIVE \cite{r3live} further improved consistency by integrating visual photometric errors and LiDAR geometric residuals into a unified ESIKF update, and introduced RGB-colored point cloud maps to enhance performance in texture-rich environments.
    
    Inspired by direct visual odometry, especially SVO \cite{svo}, FAST-LIVO \cite{fastlivo1} minimized patch-level photometric errors to reduce visual frontend computation. In the LiDAR branch,  VoxelMap \cite{voxelmap} improved registration efficiency and noise adaptability through probabilistic voxel-wise plane fitting, and was later extended into the complete LiDAR-inertial SLAM system Voxel-SLAM \cite{voxel_slam}. These ideas were integrated into FAST-LIVO2 \cite{fastlivo2}, which combines an SVO-style frontend with a VoxelMap-style backend and achieves SOTA local accuracy. However, due to the lack of global constraints, FAST-LIVO2 still accumulates drift in long-distance tasks and may fail when visual and geometric degradations occur simultaneously.
    
    \subsection{GNSS-Aided Fusion System}
    \label{subsec:gnss_related}
    To suppress accumulated drift and ensure global consistency, GNSS integration has evolved from loose coupling to tightly-coupled schemes using raw carrier-phase measurements \cite{graphgnss, tdcp_slam}.
    
    Early methods commonly adopted loose coupling. For example, LIO-SAM \cite{liosam} directly introduced GPS positions into a factor graph to constrain long-term LIO drift. However, such methods depend on receiver-level solutions and ignore correlations among raw GNSS observations, making them vulnerable to severe multipath. To better exploit raw measurements, tightly-coupled frameworks were developed. GVINS \cite{gvins} fused raw pseudorange and Doppler measurements in a nonlinear optimization framework, improving robustness, while GLIO \cite{glio} extended this idea to LiDAR systems for smooth transitions in GNSS-denied environments. Nevertheless, meter-level pseudorange noise still limits local precision.
    
    Carrier-phase-based fusion has recently attracted increasing attention for high-precision estimation. To avoid integer ambiguity resolution on mobile platforms, Time-Differenced Carrier Phase (TDCP) has been adopted. LIGO \cite{ligo_tro25} directly introduced TDCP constraints into the LIO framework, using millimeter-level relative displacement observations to suppress odometry drift. Despite these advances, existing methods remain limited in two aspects: their reliance on geometric constraints makes them vulnerable in geometry-degraded scenarios, and they lack adaptive robust mechanisms for different sensor failure modes under hybrid high-dynamic and multipath environments. Our framework builds on these tightly-coupled ideas to achieve robust mapping by intelligently fusing LIVO and GNSS.
        
\section{METHODOLOGY}
    \begin{figure*}[t]
    \centering
    \includegraphics[width=1\textwidth]{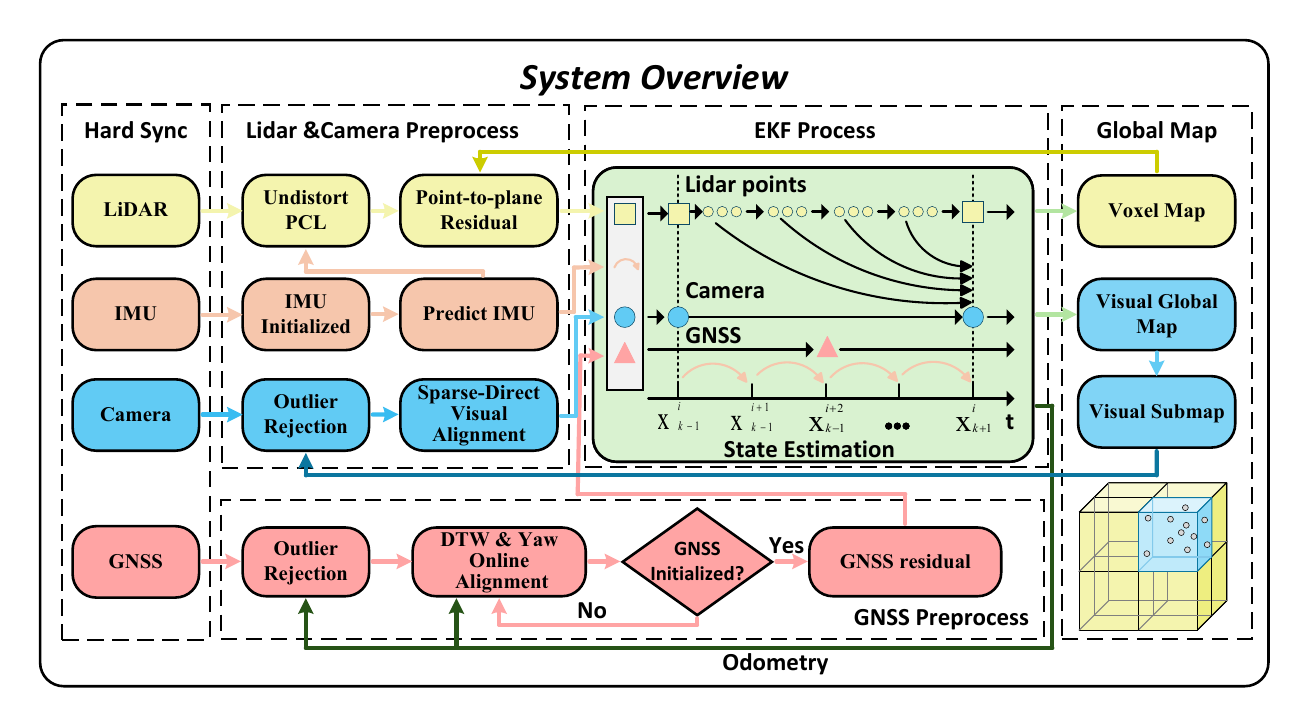}
    \caption{System overview of our proposed framework.}
    \label{System overview of FAST-LIVOTM}
    \vspace{-0.3cm}
    \end{figure*}

\subsection{System Overview\label{Overview}}

The proposed tightly-coupled multi-sensor fusion framework is built upon an Error-State Iterated Kalman Filter (ESIKF). As shown in Fig. \ref{System overview of FAST-LIVOTM}, it receives raw data from a 3D LiDAR, an RGB camera, an IMU, and a GNSS receiver. To unify spatial references, GNSS ephemeris and observations (e.g., position, velocity) are transformed from the Earth-Centered, Earth-Fixed (ECEF) frame to the local East-North-Up (ENU) frame during preprocessing, which serves as the default frame for all subsequent GNSS states.

The architecture leverages sensor complementarity: the high-frequency, robust LIVO odometry assists GNSS with online spatiotemporal synchronization, bridges signal outages, and aids quality assessment. Concurrently, GNSS carrier phase and Doppler observations eliminate LIVO's long-term drift to ensure global consistency. The processing pipeline consists of two main modules:

LIVO Update: Adopting the FAST-LIVO2 strategy, this module sequentially updates the state using camera photometric residuals and LiDAR point-to-plane geometric residuals via the ESIKF. The resulting high-frequency, sub-millimeter-accuracy local odometry serves as an initialization condition and robust geometric prior for subsequent GNSS integrity monitoring and fusion.

GNSS Update: Upon receiving GNSS data, a LIVO-assisted integrity monitor first rejects outliers. Valid measurements are then processed using DTW to align LIVO and GNSS velocity sequences. After time synchronization, a least-squares method estimates the local-to-ENU extrinsics and the GNSS lever-arm in the IMU frame. Following initialization, the ESIKF state is updated using two types of residuals:

\begin{itemize}
    \item Doppler Residual: Directly constrains the instantaneous velocity and the receiver clock drift ($\dot{\delta t}$).
    \item Time-Differenced Carrier Phase (TDCP) Residual: To obtain high-precision relative constraints, we employ a fixed-anchor strategy. Specifically, the previously updated state serves as a fixed anchor to formulate the TDCP residual. This lightweight approach successfully avoids state augmentation while fully exploiting the sub-millimeter relative precision of carrier phase observations.
\end{itemize}

\subsection{Notation and State Transition Model\label{Notation and States Transitioan Model}}

We define the full system state vector $\mathbf{x}$ in continuous time as an element of the manifold $\mathcal{M}$. The extrinsic parameters among the LiDAR, camera, and IMU are assumed to be pre-calibrated and fixed, and the lever-arm between the IMU and the GNSS antenna is modeled as a rigid transformation. To enable tightly-coupled fusion with raw GNSS measurements, the state vector is augmented to include both the IMU navigation states and the GNSS receiver clock parameters, as summarized in Table~\ref{tab:symbols}:
\begin{table}[htbp]
    \centering
    \caption{Important Notations}
    \label{tab:symbols}
    \begin{tabularx}{\columnwidth}{l >{\raggedright\arraybackslash}X}
        \toprule
        Notations  & Meaning \\
        \midrule
        ${^E(\cdot)}$ & The vector ${(\cdot)}$ in the ENU frame. \\
        ${^G(\cdot)}$ & The vector ${(\cdot)}$ in the local global frame. \\
        ${^C(\cdot)}$ & The vector ${(\cdot)}$ in the RGB camera frame.\\
        $\mathbf{x}, \widehat{\mathbf{x}}, \bar{\mathbf{x}}$ & The ground-truth, predicted, and updated estimation of $\mathbf{x}$.\\
        $\delta \mathbf{x}$ & The error state between ground-truth $\mathbf{x}$ and its estimation. \\
        ${}^I\mathbf{T}_L$ & The extrinsic of LiDAR frame w.r.t. IMU frame. \\
        ${}^I\mathbf{T}_C$ &  The extrinsic of camera frame w.r.t. IMU frame. \\
        ${}^I\mathbf{t}_r$ &  The extrinsic (lever-arm) of GNSS Receiver w.r.t. IMU. \\
        \bottomrule
    \end{tabularx}
\end{table}

In a tightly-coupled formulation, GNSS observables depend not only on the platform pose and velocity but also on receiver clock states and inter-system biases. Therefore, we augment the IMU navigation state with the receiver clock offset/drift and the multi-constellation time-bias vector, and define the continuous-time system state on $\mathcal{M}$ as
\begin{equation}
\mathbf{x} \triangleq \big[ 
    {}^G\mathbf{R}_{I}^\top~~ {}^G\mathbf{p}_{I}^\top~~ {}^G\mathbf{v}_{I}^\top~~ \mathbf{b}_{\mathbf{g}}^\top ~~ \mathbf{b}_{\mathbf{a}}^\top ~~ {}^G\mathbf{g}^\top ~~\gamma ~~ c\delta\mathbf{t}_r^\top ~~ \delta\mathbf{\dot{t}}_{r}^\top 
    \big]^\top,
\end{equation}
where $^G\mathbf{R}_{I}$, $^G\mathbf{p}_{I}$, and $^G\mathbf{v}_{I}$ describe the attitude, position, and velocity of the IMU in the local global frame, respectively, and $^G\mathbf{g}$ denotes the gravity vector. $\delta \mathbf{t}_r = [\delta t_{gr}, \delta t_{br}, \delta t_{ar}, \delta t_{or}]^\top \in \mathbb{R}^4$ is a vector representing the receiver time biases across four independent satellite systems ($gr$: GPS, $br$: BeiDou, $ar$: Galileo, $or$: GLONASS).

\subsection{Preprocess}\label{preprocess}
\subsubsection{Temporal Calibration via DTW}\label{Temporal Calibration via DTW}
 Before GNSS alignment, the LIV subsystem is initialized independently: LiDAR, camera, and IMU timestamps are hardware-synchronized, while the LiDAR-camera spatial extrinsics are calibrated using FAST-Calib \cite{fast_calib} and then kept fixed. Since the GNSS receiver and LIVO system run on independent clocks, a time offset $\delta t_{rI}$ remains between the two data streams. Given that velocity magnitude is rotation-invariant, we align the velocity sequences of LIVO and GNSS, where $v_{R} = \{ \| \mathbf{v}_{GNSS_k} \| \}$ denotes the GNSS Doppler-derived speed sequence. We employ Dynamic Time Warping (DTW) \cite{dtw_muller} to find the optimal alignment path. Compared with cross-correlation methods or online temporal calibration \cite{qin_calib}, DTW is more robust against nonlinear sampling jitter and clock drift. The resulting optimal time lag $\delta t_{rI}$ is then compensated when utilizing GNSS observations.

\subsubsection{Spatial Alignment}\label{Spatial Alignment}
After timestamp alignment, with the LIV extrinsics fixed, we use Least Squares (LS) to jointly optimize the transformation from the local global frame $\{G\}$ to the ENU frame $\{E\}$, along with the GNSS lever-arm ${}^I\mathbf{t}_{r}$. For each synchronized epoch $k$, the geometric relationship between the GNSS position observation ${}^E\mathbf{p}_{r_k}$ in ENU and the LIVO-predicted IMU position ${}^G\mathbf{p}_{I_k}$ is
\begin{equation}
\label{eq:init_ls}
{}^E\mathbf{p}_{r_k} = {}^E\mathbf{R}_G \left( {}^G\mathbf{p}_{I_k} + {}^G\mathbf{R}_{I_k} {}^I\mathbf{t}_r \right) + {}^E\mathbf{p}_G,
\end{equation}
where ${}^E\mathbf{R}_G \in SO(3)$ and ${}^E\mathbf{p}_G \in \mathbb{R}^3$ represent the rotation and translation from the local global frame to ENU. We construct the following objective function to solve for the parameter set $\mathcal{X}_{init} = \{ {}^E\mathbf{R}_G, {}^E\mathbf{p}_G, {}^I\mathbf{t}_r \}$:
\begin{equation}
\min_{\mathcal{X}_{init}} \sum_{k=1}^{N} \left\| {}^E\mathbf{p}_{r_k} - \left( {}^E\mathbf{R}_G ({}^G\mathbf{p}_{I_k} + {}^G\mathbf{R}_{I_k} {}^I\mathbf{t}_r) + {}^E\mathbf{p}_G \right) \right\|^2.
\end{equation}

This LS problem provides an accurate initial global pose and refines the coarse lever-arm extrinsics.

\section{State Estimation}\label{State Estimation}
\subsection{Propagation}\label{Propagation}
The propagation step acts as the prediction phase in the ESIKF framework, bridging the gap between discrete LiDAR/Camera/GNSS updates. The forward propagation integrates IMU measurements $\mathbf{u}_i$ from $t_{k-1}$ to $t_k$ based on a continuous-time kinematic model \cite{esikf_trawny}, assuming zero process noise $\mathbf{w}_i$ during this phase. The predicted state and covariance are denoted as $\widehat{\mathbf{x}}_k$ and $\widehat{\mathbf{P}}_k$, which serve as priors for the observation updates. Note that the GNSS clock bias term $\delta \mathbf{t}_r$ follows a constant drift-rate model. Furthermore, to compensate for motion distortion in LiDAR scans, we employ a backward propagation method to ensure all points within a scan are uniformly projected to the update time $t_k$. For brevity, the subscript $k$ is omitted in subsequent state definitions unless necessary.

\subsection{LiDAR Update}\label{Lidar Update}
For the LiDAR update, we use a point-to-plane observation model. Let ${}^L\mathbf{p}_j$ denote the undistorted point in the current scan. Using the predicted prior state $\widehat{\mathbf{x}}$, each point ${}^L\mathbf{p}_j$ is projected from the LiDAR frame $\{L\}$ to the local global frame $\{G\}$:
\begin{equation}
\label{eq:point_projection}
{}^G\hat{\mathbf{p}}_j = {}^G\hat{\mathbf{T}}_I{}^I\mathbf{T}_L{}^L\mathbf{p}_j,
\end{equation}
where ${}^G\hat{\mathbf{T}}_I$ and ${}^I\mathbf{T}_L$ represent the estimated pose and the extrinsic transformation from LiDAR to IMU, respectively.

We employ an efficient VoxelMap \cite{voxelmap} to maintain the local map. For each projected point ${}^G\hat{\mathbf{p}}_j$, we search for the nearest voxel containing a geometric plane. If a valid plane (described by normal $\mathbf{n}_j$ and center $\mathbf{q}_j$) is found, a point-to-plane residual is constructed. The observation residual $r_{\mathcal{L},j}$ is defined as the orthogonal distance from the queried point to the corresponding map plane:
\begin{equation}
0 = r_{\mathcal{L},j} (\mathbf{x}_k,{}^L\mathbf{p}_j) = \mathbf{n}_j^\top \left( {}^G\mathbf{T}_I{}^I\mathbf{T}_L{}^L\mathbf{p}_j - \mathbf{q}_j \right).
\end{equation}

For the observation function $h_l(\hat{\mathbf{x}},v_l)$, its noise term $v_l = (\delta{}^L\mathbf{p}_j,\delta{\mathbf{n}_j},\delta{\mathbf{q}_j})$ comprises raw LiDAR observation noise, normal fitting noise, and plane center point noise, respectively.

\subsection{Visual Update}\label{Visual Update}

We adopt the direct method as the visual measurement model. The direct method minimizes the photometric error (intensity difference) between corresponding patches, achieving sub-pixel accuracy and better robustness in texture-less environments \cite{svo, fastlivo1}.

Let ${}^G\mathbf{p}_i$ denote a point in the visual map associated with a reference image patch $I_r$. When a new image $I_k$ is captured at time $k$, we project the point onto the current image plane using the system state:
\begin{equation}
\label{eq:cam_projection}
\mathbf{u}_{i,k} = \pi \left( {}^C\mathbf{T}_I \left( {}^G\hat{\mathbf{T}}_I \right)^{-1} {}^G\mathbf{p}_i \right),
\end{equation}
where $\pi(\cdot)$ denotes the camera projection model, and ${}^C\mathbf{T}_I$ is the IMU-to-camera extrinsic. We define the photometric residual $r_{\mathcal{V},i}$ mapped to the small patch $\mathcal{N}$ centered at the projected pixel, requiring the intensity difference to be minimized:
\begin{align}\label{eq:visual_residual_split}
0 &= r_{\mathcal{V},i}(\mathbf{x}_k, {}^G\mathbf{p}_i) \notag\\
&= \sum_{\Delta \mathbf{u} \in \mathcal{N}} \Big\| \gamma_k I_k (\mathbf{u}_{i,k} + \Delta \mathbf{u}) - \gamma_r I_r (\mathbf{u}_{i,r} + \mathbf{A}_i^r \Delta \mathbf{u}) \Big\|,
\end{align}
where $\Delta \mathbf{u}$ is the pixel offset within the patch, $\mathbf{A}_i^r$ is an affine warp matrix compensating for scale and perspective distortion, and $\gamma$ is the exposure compensation parameter.

\subsection{GNSS Update}
\label{GNSS update}
\subsubsection{Single-Difference Carrier Phase}
\label{Single-Difference Carrier Phase}
Carrier phase measurements provide millimeter-level accuracy. Let the carrier phase observation be $\Phi_{k}^{m,j}$, where $m$ represents the constellation system, $j$ is the satellite number, and $k$ denotes the epoch. It represents the number of transmission cycles of satellite $j$ based on system $m$ with wavelength $\lambda_j$. The absolute pseudorange measurement relationship is modeled as
\begin{align}\label{eq:phase_model_meter}
\lambda_j\Phi_{k}^{m,j} =& \| {}^E\mathbf{p}_k^j - {}^E\mathbf{p}_k^r \| + \mathcal{T}_k^j - \mathcal{I}_k^j + c(\delta t_k^r - \delta t_k^j) \notag\\ 
&+ \lambda_j (\kappa_j - N_j) + \mathcal{M}_k^j + \epsilon_{\Phi},
\end{align}
where ${}^E\mathbf{p}_k^j$ and ${}^E\mathbf{p}_k^r$ denote the ENU positions of satellite $j$ and receiver antenna $r$, respectively. $\mathcal{T}_k^j$ and $\mathcal{I}_k^j$ are tropospheric and ionospheric errors. $\delta t_k^r$ and $\delta t_k^j$ are the receiver and satellite clock biases. $\kappa_j$ originates from the phase wind-up effect. $N_j$ is the unknown integer ambiguity, $\mathcal{M}_k^j$ represents the multipath effect, and $\epsilon_{\Phi}$ is the measurement noise.

Unlike pseudoranges, carrier phases cannot be directly used as absolute distance measurements due to the integer ambiguity. Thus, we construct a Time-Differenced Carrier Phase (TDCP) model, eliminating clock jumps and shared atmospheric errors via cross-epoch single differencing \cite{ rtklib}. Assuming no cycle slip occurs, the constant integer ambiguity is completely canceled out.

The single-differenced observation equation is derived as
\begin{align}\label{eq:sd_derivation}
\lambda_j(\Phi_{k+1}^{m,j} - \Phi_{k}^{m,j}) = & \| {}^E\mathbf{p}_{k+1}^j - {}^E\mathbf{p}_{k+1}^r \| - \| {}^E\mathbf{p}_k^j - {}^E\mathbf{p}_k^r \| \notag\\
&+ (\mathcal{T}_{k+1}^j - \mathcal{T}_k^j) - (\mathcal{I}_{k+1}^j - \mathcal{I}_k^j) \notag\\
&+ c(\delta t_{k+1}^r - \delta t_k^r)- c(\delta t_{k+1}^j - \delta t_k^j) \notag\\
&+ (\mathcal{M}_{k+1}^j - \mathcal{M}_k^j) + \Delta \epsilon_{\Phi}.
\end{align}

For high-sampling-rate GNSS, the spatial variation of ionospheric and tropospheric delays over a short interval $\Delta{t}$ is negligible compared to carrier phase noise. Hence, we assume that
\begin{equation}
\Delta \mathcal{T}_k^j \approx 0, \quad \Delta \mathcal{I}_k^j \approx 0.
\end{equation}

The specific calculation mechanism of the TDCP residual is visually summarized in the following Fig. \ref{fig:tdcp_model}:
\begin{figure}[htbp]
\centering
\includegraphics[width=\linewidth]{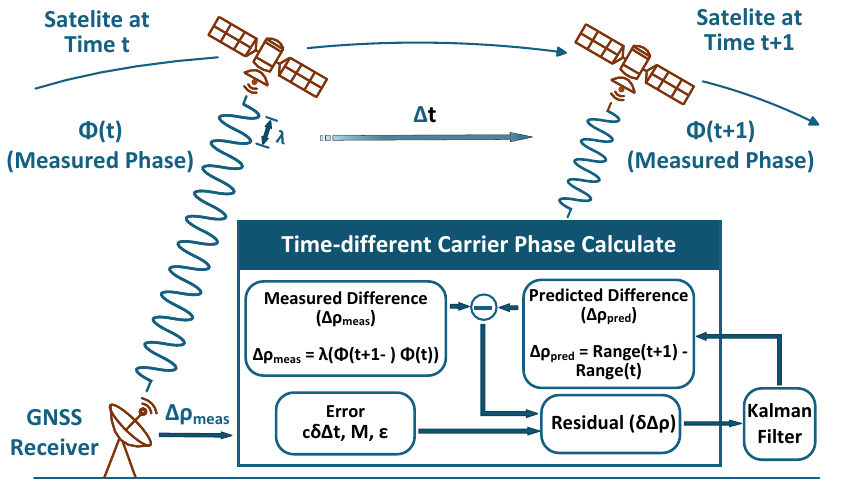}
\caption{Schematic of the Time-Differenced Carrier Phase (TDCP) model. Consecutive epoch differencing provides precise relative position constraints ($\Delta\rho$) by eliminating integer ambiguities and mitigating atmospheric delays.}
\label{fig:tdcp_model}
\end{figure}

Unlike atmospheric errors, the variation in multipath error $\Delta \mathcal{M}$ in complex scenarios (e.g., urban canyons) is severe and unpredictable. Grouping unmodeled multipath and noise into the residual, the actual TDCP residual equation for the ESIKF update is derived as
\begin{align}\label{eq:tdcp_final_compact}
    r_{\Delta \Phi}^{m,j} &= 
    \Big(\lambda_j (\Phi_{k+1}^{m,j} - \Phi_{k}^{m,j}) + c (\delta t_{k+1}^j - \delta t_k^j) \Big) \notag\\
    &\quad - \Big(\rho_{k+1}^{r,j} - \rho_{k}^{r,j} + c (\delta t_{k+1}^r - \delta t_k^r)\Big),
\end{align}
where $\rho_{\tau}^{r,j}$ is defined as the geometric distance between the phase center of the satellite and receiver antenna at time $\tau \in \{k, k+1\}$:
\begin{equation*}
    \rho_{\tau}^{r,j} = \left\| {}^E\mathbf{p}_\tau^j - \left( {}^E\mathbf{R}_G \left( {}^G\mathbf{p}_{I_\tau} + {}^G\mathbf{R}_{I_\tau} {}^I\mathbf{t}_r \right) + {}^E\mathbf{p}_G \right) \right\|.
\end{equation*}

To constrain relative motion without state augmentation, we adopt an "anchor-based" approach, treating the previous posterior state $\hat{\mathcal{X}}_{k} = \{ {}^G\hat{\mathbf{p}}_{I_k}, {}^G\hat{\mathbf{R}}_{I_k}, \delta \hat{\mathbf{t}}_k^r \}$ as a fixed anchor. Thus, epoch $k$ terms ($\rho_k^{r,j}$, $\delta t_k^r$) in Eq. (\ref{eq:tdcp_final_compact}) become constants. By linearizing the observation model solely against the current error state $\delta \mathbf{x}_{k+1}$, TDCP effectively acts as a high-precision odometry constraint. Furthermore, uneliminated multipath errors $\Delta \mathcal{M}$ induce anomalous peaks in $r_{\Delta \Phi}^{m,j}$, facilitating LIVO-prior-based outlier rejection.

\subsubsection{Doppler Frequency Measurement Model}
\label{subsec:doppler_model}

The raw Doppler shift observation $\mathcal{D}_k^j$ (Hz) provides an instantaneous measurement of the relative velocity between the satellite and receiver, modeled as
\begin{align}\label{eq:doppler_raw}
\mathcal{D}_k^j = &-\frac{1}{\lambda_j} (\mathbf{u}_k^j)^\top \left( {}^E\mathbf{v}_k^j - {}^E\mathbf{v}_k^r \right) \notag\\&- \frac{c}{\lambda_j} (\dot{\delta t}_k^r - \dot{\delta t}_k^j) + \epsilon_{\mathcal{D}},
\end{align}
where $\mathbf{u}_k^j$ is the line-of-sight unit vector. ${}^E\mathbf{v}_k^j$ and ${}^E\mathbf{v}_k^r$ represent the velocity vectors of the satellite and receiver in ENU. $\dot{\delta t}_k^r$ and $\dot{\delta t}_k^j$ are the clock drift rates.

By projecting the estimated velocity into the ENU frame using ${}^E\mathbf{R}_G$, the Doppler residual $r_{\mathcal{D}}^{m,j}$ is defined as
\begin{align}
\label{eq:doppler_residual_clean}
    r_{\mathcal{D}}^{m,j} =& \left( -\lambda_j \mathcal{D}_k^j + c \cdot \dot{\delta t}_k^j \right) \notag\\
    &- \left[ 
    (\mathbf{u}_k^j)^\top \left( 
        {}^E\mathbf{v}_k^j - {}^E\mathbf{R}_G{}^G\hat{\mathbf{v}}_{k}^r 
    \right) 
    + c \cdot \hat{\dot{\delta t}}_k^r 
    \right].
\end{align}

Considering the lever-arm effect from IMU to the receiver antenna, the antenna velocity is compensated as 
\begin{align*}
{}^G\hat{\mathbf{v}}_{k}^r ={}^G\hat{\mathbf{v}}_{I_k} + {}^G\hat{\mathbf{R}}_{I_k}\left( {}^I\hat{\boldsymbol{\omega}}_k \times {}^I\mathbf{t}_{r} \right).
\end{align*}

\subsection{Hierarchical Integrity Monitoring and Adaptive Fusion}
\label{Hierarchical Integrity Monitoring}

Since ESIKF is extremely sensitive to anomalous observations, a strict outlier rejection strategy is the prerequisite for ensuring global consistency. The strategy is divided into two hierarchical parts:

1) GNSS Internal Integrity Check (RAIM):
Before introducing GNSS observations, a consistency check relying solely on internal receiver information is performed. Signals with low Carrier-to-Noise density (C/N0 $<$35 dB-Hz) or low elevation angles ($<15^{\circ}$) are physically filtered out. Subsequently, Single Point Positioning (SPP) using Least Squares is performed with the remaining healthy satellites. Any satellite with pseudorange residuals significantly deviating from the overall distribution is forcefully excluded.

2) Degeneracy-Dependent Adaptive Rejection:
We extract the minimum eigenvalue $\lambda_{min}$ of the Hessian matrix during the LIVO update as a reliable measure of local geometric/textural constraint strength. To avoid high-frequency mode switching caused by sensor noise (the ping-pong effect), the system maintains a sliding window to calculate the smoothed mean $\bar{\lambda}_{min}$. The system adaptively switches between two modes based on the degradation threshold $\lambda_{th}$:

\textit{Mode A: LIVO Well-Conditioned ($\bar{\lambda}_{min} \ge \lambda_{th}$).} The LIVO prior is highly reliable. We apply a rigorous Chi-Square Test to the GNSS observations $\mathbf{z}_j$ that passed RAIM:
\begin{equation}
\mathbf{r}_j^\top (\mathbf{H}_j \widehat{\mathbf{P}}_k \mathbf{H}_j^\top + \mathbf{V}_j)^{-1} \mathbf{r}_j < \chi^2_{th},
\end{equation}
where $\mathbf{r}_j$ is the residual between the GNSS measurement and the LIVO prediction. $\widehat{\mathbf{P}}_k$ and $\mathbf{V}_j$ are the prior and observation covariances. This mode leverages the high-confidence prior to effectively eliminate subtle multipath deviations, ensuring trajectory smoothness.

\textit{Mode B: LIVO Degraded ($\bar{\lambda}_{min} < \lambda_{th}$).} When encountering feature degradation (e.g., textureless long corridors), the prior covariance $\widehat{\mathbf{P}}_k$ inflates. The system adaptively aborts the strict prior-based test, directly adopting all GNSS observations that passed the RAIM check. Under extreme conditions lacking local constraints, this mechanism fully trusts the absolute positioning capability of GNSS, effectively preventing catastrophic drift.

\begin{table*}[htbp]
    \centering
    \caption{Quantitative Comparison of ATE RMSE (m) on the M3DGR Dataset Across Different Evaluated Algorithms.}
    \label{tab:m3dgr_results_transposed}
    \resizebox{\textwidth}{!}{
    \begin{tabular}{lccccccc}
        \toprule
        \multirow{2}{*}{\textbf{Algorithm}} & \multicolumn{5}{c}{\textbf{Open Sky Scenarios}}  & \multicolumn{2}{c}{\textbf{GNSS Denied Scenarios}} \\
        \cmidrule(lr){2-6}  \cmidrule(lr){7-8}
         & \textit{Varying-illu03} & \textit{Varying-illu04} & \textit{Varying-illu05} & \textit{Outdoor01} & \textit{Outdoor04} & \textit{Dark01} & \textit{Dark02} \\
        \midrule
        RTKLIB       & 7.50 & 7.97 & 5.62 & 2.08 & 3.45 & 16.39 & 40.88  \\
        LIO-SAM-GPS  & 1.93 & 1.01 & 4.83 & 0.45 & 0.58 & 1.85  & 3.62   \\
        FAST-LIVO2   & 1.44 & 0.35 & 0.16 & 0.27 & 0.25 & \textbf{0.31} & \textbf{0.18}  \\
        LIGO         & 0.98 & 0.26 & 0.28 & 0.33 & 0.36 & 2.45  & 4.12   \\
        Ours w/o Rejection & 1.05 & 0.28 & 0.13 & 0.18 & 0.20 & -  & -   \\
        Ours         & \textbf{0.23} & \textbf{0.19} & \textbf{0.09} & \textbf{0.11} & \textbf{0.13} & 0.34  & 0.27   \\
        \bottomrule
    \end{tabular}
    }
\end{table*}

\section{EXPERIMENTAL RESULTS}
\label{EXPERIMENT RESULT}

\subsection{Evaluation on Benchmark Dataset (M3DGR)}
\label{Evaluation on M3DGR}

We conducted standardized quantitative comparisons on the public M3DGR dataset \cite{m3dgr}, which provides RTK ground truth in diverse urban and forest scenarios. Although MARS-LVIG \cite{marslvig} offers large-scale aerial LiDAR-visual-inertial-GNSS data, it was excluded because LIO-SAM could not be reliably initialized on it, preventing fair comparison. We benchmarked our method against SOTA open-source systems, including LIO-SAM \cite{liosam}, FAST-LIVO2 \cite{fastlivo2}, and LIGO \cite{ligo_tro25}. An ablation variant, Ours w/o Rejection, was also evaluated by disabling the adaptive robust module and fusing all pre-screened GNSS observations.

Table \ref{tab:m3dgr_results_transposed} summarizes the quantitative results, leading to the following observations:

\subsubsection{Performance in Open-Sky Environments}
In favorable GNSS sequences (e.g., \textit{Outdoor\_01}, \textit{Outdoor\_04}), our complete method (Ours) achieves the highest accuracy. 
Benefiting from the tight coupling of global GNSS and local LIVO, the system effectively mitigates long-distance cumulative drift. Notably, although the ablation version (Ours w/o Rejection) outperforms pure FAST-LIVO2 due to GNSS integration, it remains slightly inferior to our complete version. This indicates that even in open skies, occasional weak multipath noise affects high-precision estimation. Our robust mechanism effectively filters these hidden errors to extract ultimate accuracy.

\subsubsection{Performance in GNSS-Degraded Environments}
In GNSS-degraded environments (e.g., \textit{Dark01}, \textit{Dark02}), satellite signals suffer severe occlusion, causing extreme multipath errors.

Under such harsh conditions, a critical finding is that the ablation version (Ours w/o Rejection) directly suffers from trajectory divergence and system failure (marked as '-' in the table). Because our system adopts a highly tightly-coupled ESIKF architecture, unisolated massive GNSS errors directly contaminate the error states and covariance matrix, leading to irreversible filter divergence. In contrast, looser-coupled systems like LIO-SAM and LIGO generate meter-level errors but avoid complete failure. This powerfully demonstrates that in a deeply tightly-coupled framework, our degeneracy-aware dual-mode outlier rejection is the fundamental cornerstone of system survivability and robustness.

Meanwhile, our complete system (Ours) exhibits extreme robustness here. Its accuracy is only marginally second to pure FAST-LIVO2 (a centimeter-level difference)---which is unaffected by external signals—but far surpasses LIO-SAM and LIGO. This minor error penalty is the ultimate trade-off of the fusion mechanism. By sacrificing negligible accuracy in denied environments, our system gains significant global accuracy improvements in open areas and successfully prevents tightly-coupled system crashes, achieving optimal comprehensive performance across all scenarios.

\subsection{Evaluation on Private Dataset}
\label{Evaluation on Private Dataest}

\subsubsection{Experimental Platform}
\label{Experimental Hardware Platform}

\begin{figure}[htbp]
  \centering
    \centering
    \includegraphics[width=\linewidth]{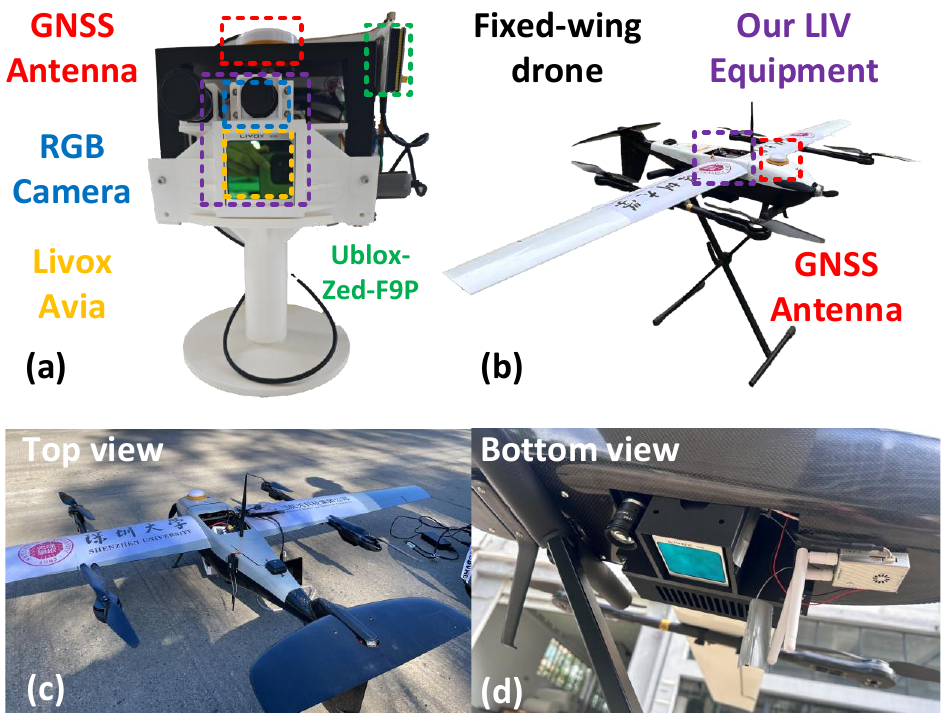}
    \caption{ Our platform with hardware synchronization for data acquisition. (a) Handheld system, (b) Fixed-wing drone setup, (c) Drone top view, (d) Drone bottom view.}
    \label{fig:hardware_setup}
\end{figure}

We constructed a modular multi-sensor data acquisition platform, as shown in Fig. \ref{fig:hardware_setup}, integrating a Livox Avia solid-state LiDAR, an industrial-grade global shutter RGB camera, and a Ublox ZED-F9P GNSS receiver. To ensure temporal consistency, hardware-triggered synchronization was implemented via a microcontroller generating PPS signals. The sensor suite was configured in two forms: handheld (indoor-outdoor transitions) and mounted on a high-speed fixed-wing UAV (large-scale dynamic mapping).

\subsubsection{High-Precision Mapping in Large-Scale Highly Dynamic Environments}
\label{High-Precision Mapping in Large-Scale High-Dynamic Environments}

\begin{figure}[htbp]
    \centering
    \includegraphics[width=\linewidth]{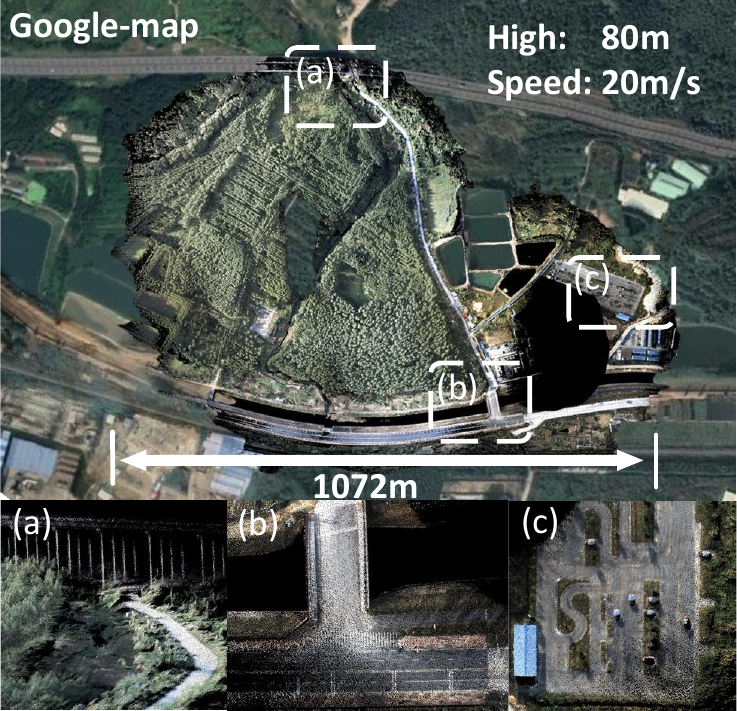}
    \caption{Trajectory overview of the large-scale fixed-wing UAV experiment overlaid on Google Maps. The path segments (a), (b), and (c) correspond to challenging maneuvering phases.}
    \label{fig:global_mapping}
\end{figure}

To verify accuracy and consistency, long-distance flight experiments were conducted. The UAV cruised at 75 m altitude with a high speed of 20 m/s, which poses severe challenges (sparse texture and motion blur). As seen in Fig. \ref{fig:global_mapping}, our system maintained stability during long straight flights and sharp maneuvers.

\begin{figure}[htbp]
    \centering
    \includegraphics[width=\linewidth]{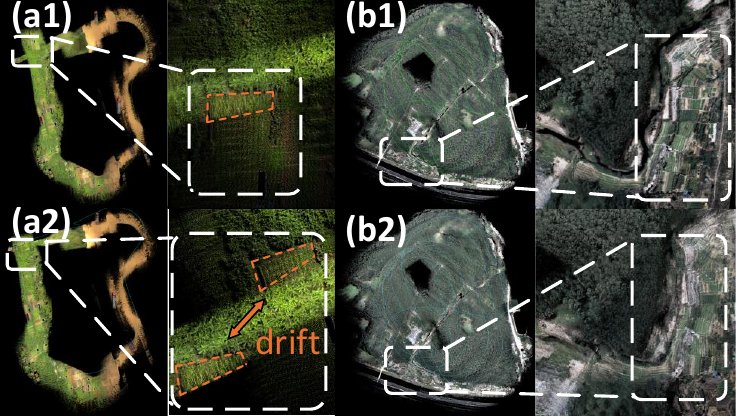}
    \caption{Qualitative comparison of global point cloud maps. (a1, b1) Our method; (a2, b2) FAST-LIVO2.}
    \label{fig:mapping_compare}
\end{figure}

A qualitative comparison of the generated global point clouds is presented in Fig. \ref{fig:mapping_compare}. Specifically, FAST-LIVO2 suffers from severe accumulated drift, causing significant structural misalignment (Fig. \ref{fig:mapping_compare} (a2)). In contrast, our method effectively eliminates this drift via tightly-coupled GNSS constraints, maintaining strict geometric alignment (Fig. \ref{fig:mapping_compare} (a1)). Furthermore, while FAST-LIVO2 exhibits obvious map ghosting and layering in complex overlapping areas (Fig. \ref{fig:mapping_compare} (b2)), our system successfully suppresses these errors, yielding a sharp, globally consistent point cloud (Fig. \ref{fig:mapping_compare} (b1)).

\begin{figure}[htbp]
    \centering
    \includegraphics[width=\linewidth]{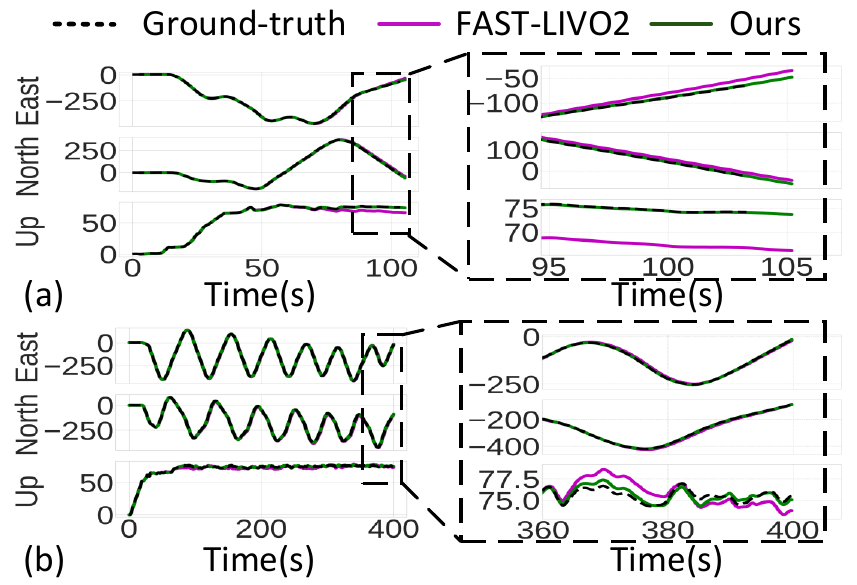}
    \caption{Trajectory evaluation in the large-scale flight. (a) and (b) show ENU position estimates across different phases. Our method (green) tightly aligns with the RTK ground truth (dashed), outperforming FAST-LIVO2 (grey).}
    \label{fig:evo_traj}
\end{figure}

To further validate this improvement, Fig. \ref{fig:evo_traj} evaluates the estimated trajectories. It demonstrates that our method tightly aligns with the RTK ground truth, effectively preventing the severe Z-axis drift observed in pure LIVO.

\subsubsection{Robustness in Handheld Hybrid Scenarios}
\begin{figure}[htbp]
    \centering
    \includegraphics[width=\linewidth]{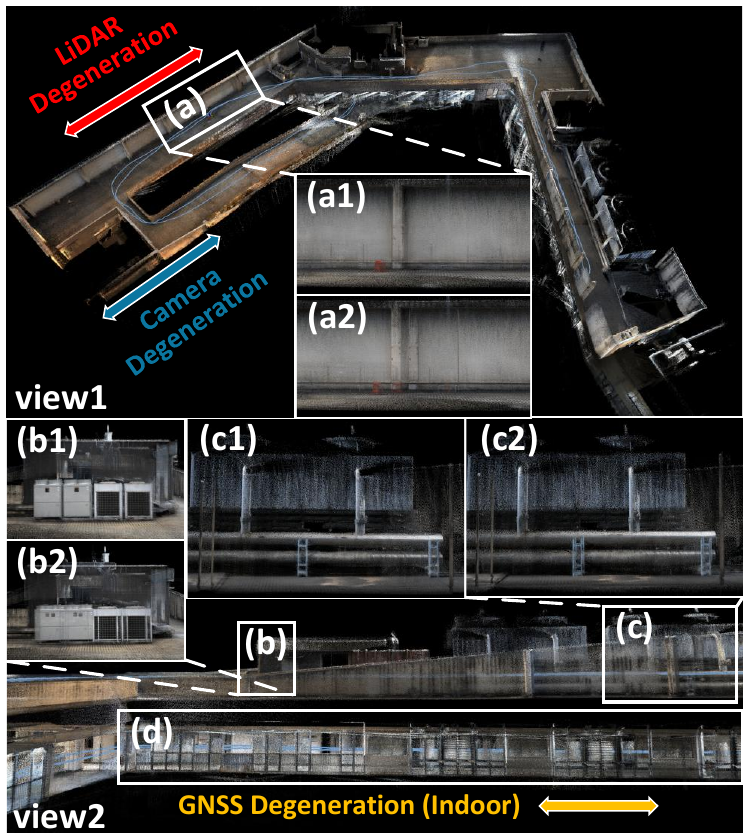}
    \caption{Mapping results generated in complex LIDAR, Camera, and GNSS degenerated scenes.}
    \label{fig:mapping_compare2}
\end{figure}

To evaluate robustness in complex environments, we collected a handheld dataset involving indoor-outdoor transitions (Fig. \ref{fig:mapping_compare2}).

1) Indoor GNSS Degraded Scenario: In the long indoor corridor (see region (d) in Fig. \ref{fig:mapping_compare2}, View 2), GNSS signals were severely occluded. Traditional schemes diverge easily here, but our strict outlier rejection strategy isolated unreliable GNSS signals. The resulting map was as flat and clear as the optimal pure odometry FAST-LIVO2.

2) Outdoor LIVO Degraded Scenario: In regions (a)-(c) of View 1 (Fig. \ref{fig:mapping_compare2}), the LiDAR lacked geometric constraints, and the camera faced texture-less walls and sudden illumination changes. FAST-LIVO2 suffered from cumulative drift, leading to map ghosting. In contrast, our system seamlessly fused the healthy GNSS signals available outdoors, rectifying the odometry drift and generating sharp boundaries, significantly outperforming pure LIVO.

\subsubsection{Real-Time Performance}
Benefiting from the efficient FAST-LIVO2 on resource-constrained platforms \cite{fastlivo2_arm} and our lightweight GNSS observation models, our system possesses excellent real-time capabilities. Table \ref{tab:time_consumption} outlines the average processing time per frame.

\begin{table}[htbp]
    \centering
    \caption{Average Time Consumption of Each Module}
    \label{tab:time_consumption}
    \begin{tabular}{lc}
        \toprule
        \textbf{Module / Operation} & \textbf{Time (ms)} \\
        \midrule
        Base LIVO LiDAR Update  & 23.36 \\
        Base LIVO Visual Update & 2.64 \\
        \midrule
        \textbf{Base LIVO Subtotal} & \textbf{25.99} \\
        \midrule
        GNSS Preprocessing (DTW \& Alignment) & 4.15 \\
        Integrity Monitoring (RAIM \& $\chi^2$ Test) & 3.42 \\
        GNSS State Update (Doppler \& TDCP) & 4.21 \\
        \midrule
        \textbf{Total GNSS Overhead} & \textbf{11.78} \\
        \midrule
        \textbf{Total System Processing Time} & \textbf{37.77} \\
        \bottomrule
    \end{tabular}
\end{table}

As detailed in Table \ref{tab:time_consumption}, the reported FAST-LIVO2-RC baseline requires 25.99 ms per frame for LIVO-only processing. The complete GNSS integration, including DTW alignment, integrity monitoring, and ESIKF updates, introduces only 11.78 ms overhead due to the fixed-anchor TDCP design. The total processing time of 37.77 ms remains well below the 100 ms budget of a standard 10 Hz LiDAR, satisfying real-time requirements while leaving computational headroom for downstream planning and control.

\section{CONCLUSION}
\label{sec:conclusion}

This paper presents a tightly-coupled LiDAR-Inertial-Visual-GNSS fusion system for highly dynamic hybrid scenarios. By combining DTW-based alignment, TDCP/Doppler observations, and degeneracy-dependent outlier rejection, the system suppresses long-term drift and improves robustness in degraded environments. Experiments on UAV and handheld datasets validate its robustness and global consistency.

\end{document}